\documentclass[10pt,twocolumn,letterpaper]{article}

\usepackage{iccv}
\usepackage{times}
\usepackage{epsfig}
\usepackage{graphicx}
\usepackage{amsmath}
\usepackage{amssymb}

\usepackage{booktabs}       % professional-quality tables
\usepackage{multirow}

\usepackage{nicefrac}       % compact symbols for 1/2, etc.

\usepackage{xcolor}         % colors
 \usepackage{float} 

\usepackage[breaklinks=true,bookmarks=false]{hyperref}

\usepackage[capitalize]{cleveref}
\crefname{section}{Sec.}{Secs.}
\Crefname{section}{Section}{Sections}
\Crefname{table}{Table}{Tables}
\crefname{table}{Tab.}{Tabs.}

\iccvfinalcopy % *** Uncomment this line for the final submission

\def\iccvPaperID{****} % *** Enter the ICCV Paper ID here

\ificcvfinal\fi

\begin{document}

%%%%%%%%% TITLE
\title{Learnt Contrastive Concept Embeddings for Sign Recognition}
\author{Ryan Wong$^1$, Necati Cihan Camgoz$^2$, Richard Bowden$^1$\\
%Centre for Vision Speech and Signal Processing, 
$^1$University of Surrey, United Kingdom\\
$^2$Meta Reality Labs, Switzerland\\
{\tt\small \{r.wong, r.bowden\}@surrey.ac.uk, neccam@meta.com}
}

\maketitle
\ificcvfinal\thispagestyle{empty}\fi

%%%%%%%%% ABSTRACT
\begin{abstract}
In natural language processing (NLP) of spoken languages, word embeddings have been shown to be a useful method to encode the meaning of words. 
Sign languages are visual languages, which require sign embeddings to capture the visual and linguistic semantics of sign.

Unlike many common approaches to Sign Recognition, we focus on explicitly creating sign embeddings that bridge the gap between sign language and spoken language.
We propose a learning framework to derive LCC (Learnt Contrastive Concept) embeddings for sign language, a weakly supervised contrastive approach to learning sign embeddings. We train a vocabulary of embeddings that are based on the linguistic labels for sign video. 
Additionally, we develop a conceptual similarity loss which is able to utilise word embeddings from NLP methods to create sign embeddings that have better sign language to spoken language correspondence.
These learnt representations allow the model to automatically localise the sign in time. 

Our approach achieves state-of-the-art keypoint-based sign recognition performance on the WLASL and BOBSL datasets.

\end{abstract}

%%%%%%%%% BODY TEXT
\section{Introduction}
\label{sec:intro}

Sign languages are visual languages used by Deaf communities around the world. Each country will typically have its own sign language which differs in their vocabulary. While the vocabulary of lexicons will vary, all sign languages share common attributes in terms of their use of hand shape, motion, 3D space, body posture, facial expression and mouthings in order to communicate. We can think of these as different channels of information.

The visual complexities of sign language make automatic Sign Language Translation (SLT) a challenging task for both natural language processing (NLP) and computer vision.
Many methods have been proposed to break the SLT problem down into simpler tasks. These tasks include Isolated Sign Recognition, where the goal is to identify a single sign in a given temporal window or Sign Spotting, which is the task of identifying and temporally locating signs within a continuous sequence. 
All these tasks are used to help solve the underlying SLT problem of translating a sign video to the spoken language equivalent. 

Many approaches tackle sign recognition as an extension to gesture recognition by making use of gesture recognition models  \cite{albanie2020bsl,albanie2021bbc,jiang2021skeleton,li2020word}.
While these approaches have been shown to be successful at classification, for the datasets they are trained on, they do not explicitly create a sign representation associated for the sign label. 
We, therefore, propose creating learnable sign embeddings which are associated with signs from the target vocabulary, similar to word embeddings used in spoken language NLP.
Signs with similar meaning or context are usually visually similar signs \cite{stokoe1980sign}. Motivated by this, we leverage word embeddings from NLP to learn better sign embeddings.
Word embeddings, such as Word2Vec \cite{mikolov2013efficient}, fastText \cite{bojanowski2017enriching} and GloVe \cite{pennington2014glove}, are useful in NLP as the vector representations capture semantic similarities between words, where words with similar meanings are closer in the embedding space.
In this paper, we guide the learning of sign embeddings by bringing together signs with similar meanings in the embedding space.

We summarise our contributions as follows: 
(1) We introduce a Learnt Contrastive Concept (LCC) embedding framework, which is a weakly supervised contrastive learning approach to explicitly learn sign embeddings with capabilities for automatic sign localisation.
(2) We propose a novel method to integrate spoken language word embeddings with sign labels to produce a representation that has better sign-to-spoken language correspondence. 
(3) Our loss function is able to improve skeletal and RGB-based model performance compared to cross entropy loss used in past approaches. 
(4) Our approach is able to outperform previous skeleton-based state-of-the-art models on the WLASL and BOBSL recognition tasks.

\section{Related Work}

The advancements in computational power and deep learning have shifted the focus of sign recognition from hand-crafted features to data-driven methods.
Various approaches to solving sign recognition have been explored over the years, such as breaking the problem into sub-problems by creating hand and mouthing shapes to be used for sign related tasks \cite{cihan2017subunets, koller2015deep}. These hand and mouthing shapes have been shown to be useful for both sign recognition and SLT \cite{camgoz2020multi,camgoz2020sign}, but they require labels to be manually annotated at the frame level or learnt with specialised sign classification models \cite{koller2019weakly}. One possible source of phonetic representation comes from the linguistic annotation in the form of HamNoSys or SignWriting \cite{hanke2004hamnosys,sutton2009signwriting}. Such annotation systems provide detailed transcription for sign languages but require expert annotators to transcribe the video which is an expensive and time consuming task.

The development of large scale sign recognition datasets has allowed the exploration of deep learning based approaches. 
Continuous Sign Recognition datasets were initially created such as RWTH-PHOENIX-Weather-2014 \cite{forster2014extensions} and RWTH-PHOENIX-Weather-2014-T \cite{camgoz2018neural} for the task of predicting all signs in a given sign video. Many approaches tackle this task with models trained with Connectionist Temporal Classification (CTC) loss \cite{camgoz2020sign,hao2021self,min2021visual}.

In this paper we focus on the sign recognition task. The AUTSL \cite{sincan2020autsl}, WLASL \cite{li2020word} and MSASL \cite{joze2018ms} datasets have been developed for sign recognition in an isolated setting, as well as developments of sign recognition from co-articulated videos such as BOBSL \cite{albanie2021bbc}. 
To tackle the sign recognition tasks on these datasets, action recognition models are used as inspiration. The inflated 3D ConvNet (I3D) \cite{carreira2017quo} and ResNet2+1D \cite{tran2018closer} used in human action recognition has proven useful for transfer learning from large action recognition datasets to smaller isolated sign recognition datasets \cite{albanie2020bsl,albanie2021bbc,jiang2021sign}. 
The pretrained sign models are used as feature extractors for the sign language translation task, reducing the RGB frames to a lower dimensional vector \cite{albanie2021bbc,camgoz2020multi}.

Such methods that use raw video as input assume the model will learn the person independent features. An alternative approach is to use skeleton-based models to reduce the impact of background noise and the person's appearance. 
Most skeleton-based approaches make use of pose estimators like OpenPose or MediaPipe to detect and extract human body, hand and facial keypoints \cite{cao2019openpose,lugaresi2019mediapipe,zhang2020mediapipe}. 
Skeletal inputs have been shown to be useful in Action Recognition tasks using Spatial-Temporal Graph Convolutional Networks to automatically learn the spatial and temporal patterns from the data \cite{yan2018spatial}. Jiang \etal proposed the Sign Language Graph Convolution Network (SL-GCN) for sign recognition \cite{jiang2021skeleton}. One of the drawbacks to skeletal-based models is the requirement for accurate keypoint detection in the presence of motion, making sign recognition highly reliant on the pose estimator's accuracy \cite{moryossef2021evaluating}. 

Learning sign language specific representations is currently an under explored topic but learning representations via contrastive learning is popular in computer vision \cite{chen2020improved, khosla2020supervised,schroff2015facenet}. These approaches attempt to learn embeddings where visually similar samples are close and dissimilar ones are far away in the embedding space. 
% However, such approaches required large datasets and the classes in these datasets must have different visual features.
Sign language is a fine grain classification task which requires a subtle understanding of the small differences between pose, hand shape, motion and mouthings/face gestures.
The multiple channels convey complementary information and machine learning models need to learn how these channels interact to understand the sign content.
This is especially challenging since models need to learn features that are signer independent across datasets that may have a fairly low number of signers.
Bilge \etal look into creating sign representations using zero-shot sign recognition approaches but require large quantities of annotations of textual descriptions of the motion and body pose on these datasets \cite{bilge2022towards}. 
Albanie \etal have demonstrated how I3D feature representations are useful for automatic dense annotations for large-vocabulary sign language videos using cosine similarity to create a dataset of dense sign spottings \cite{momeni2022automatic}.

While the above approaches have focused on visual similarities in sign, Dafnis \etal highlight the importance of linguistics priors \cite{dafnis2022bidirectional}. They modify the labels of WLASL sign recognition dataset to provide a 1-1 correspondence between the signs and glosses by comparing the labels of videos from the WLASL dataset to videos from a sign bank. They additionally use an external sign dataset to explicitly learn to detect the start and end frames of the signs as a preprocessing step before sign recognition. In this work, we demonstrate an approach that eliminates the need for preprocessing and automatically localises the signs while classifying them.

Since sign languages are visual languages, we take a different approach and look towards NLP to tackle sign recognition. Word embeddings have been found to be useful to represent spoken words as vectors for text analysis, where words with similar embeddings have similar meaning. Word2Vec \cite{mikolov2013efficient} and fastText \cite{bojanowski2017enriching} are examples of such methods to train meaningful embedding vectors.  
Motivated by this, we aim to learn embeddings for the sign recognition task which brings signs that are visually and linguistically similar closer together in the embedding space.

\section{Method}

\begin{figure*}[ht]
    \centering
    \includegraphics[width=\textwidth]{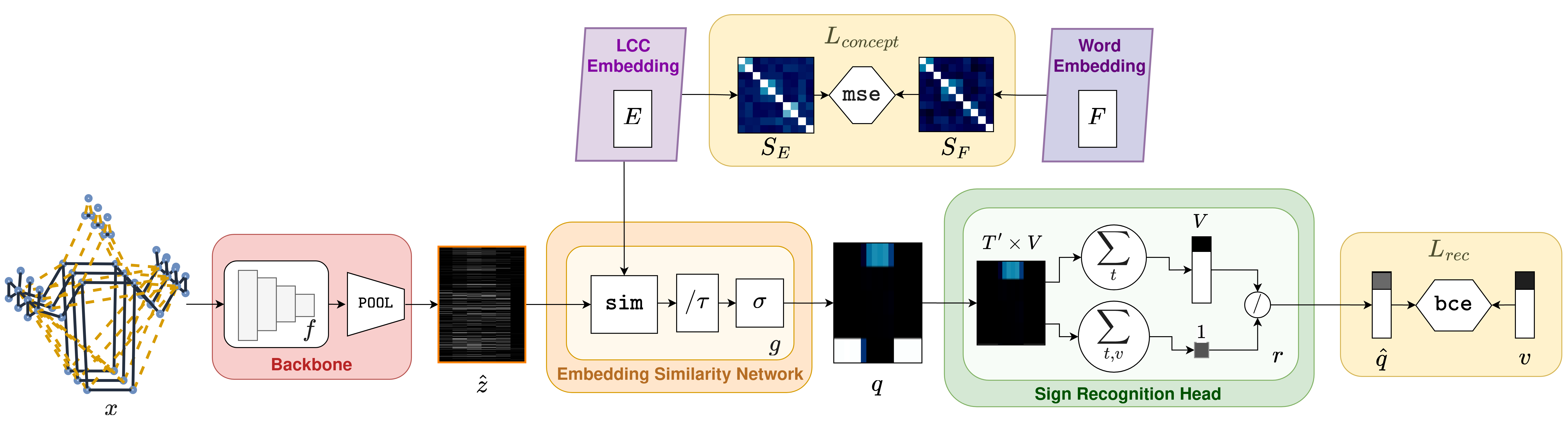}
    \caption{Overview of our proposed Learnt Contrastive Concept (LCC) embedding framework.}
    \label{fig:spatialtemporalmodel}
\end{figure*}

As shown in Figure~\ref{fig:spatialtemporalmodel}, our proposed framework has three core components: an \textit{LCC Embedding}, an \textit{Embedding Similarity Network} and a \textit{Sign Recognition Head} for the sign recognition task.
The LCC embedding captures the sign information which has similar functionality to word embeddings in NLP, where similar signs will have similar representations. The embedding similarity network and sign recognition head are used to automatically localise and predict the sign within the sequence.
Unlike previous approaches, which make use of only the sign labels to indirectly learn sign representations from visual features, we explicitly learn the sign representations and incorporate linguistic knowledge to guide the model to learn visual-linguistic representations.

In the following section, we describe the architectural changes to support our framework. Then we elaborate on the incorporation of our visual-linguistic loss functions namely, a contrastive recognition loss and a conceptual similarity loss.

\subsection{Model Architecture}

\paragraph{Backbone $f(\cdot)$:}

Our backbone uses human pose keypoints as input (See Section~\ref{sec:multi_channel} for further details about the input structure).
Given a sign sequence $x\in\mathbb{R}^{T\times D \times N}$ of $T$ length with $N$ nodes and feature channels $D$, we want to maximise the accuracy of our model to predict the target label from a vocabulary $V$. 
The most common approach to train such a recognition model is:
\begin{equation}
    \label{eq:normal}
    \hat{y} = \texttt{FC}(\texttt{POOL$_{\texttt{global}}$}(f(x)))
\end{equation}
where $\hat{y}\in\mathbb{R}^{V}$ is the logit class prediction. $\texttt{FC}$ is a fully connected layer and $\texttt{POOL}_{\texttt{global}}$ is the spatio-temporal global average pooling layer which takes the output representation from the model $f(x)=z\in\mathbb{R}^{\frac{T}{\sigma_{t}} \times C \times \frac{N}{\sigma_{n}}}$ to a vector $\hat{z} \in \mathbb{R}^{C}$. $\sigma_t$ and $\sigma_n$ are the dimension reduction factors from the backbone network for $T$ and $N$ respectively.

Instead of spatio-temporal global average pooling, we use spatial global average pooling such that $\hat z\in\mathbb{R}^{\frac{T}{\sigma_{t}} \times C}$, which allows our model to make more fine-grained predictions. For simplicity we will refer to $\frac{T}{\sigma_{t}}$ as $T'$.

\paragraph{LCC Embedding $E$:}

To disentangle the embedding representation from the model representation ($\hat{z}$), we introduce LCC embeddings $E\in\mathbb{R}^{C\times V' \times M}$ as the sign embedding representation, where $V'$ is the selected vocabulary size and $M$ is the number of variations associated with each item in the embedding vocabulary. We select $V'$ where $V<V'$. The motivation behind this choice is that not all representations within the sign sequence should be associated with an embedding in the given target vocabulary (from index $1$ to $V$) as they may be \textit{background}, such as signers in their resting pose, \textit{transitions}, or \textit{out of vocabulary signs}.

\paragraph{Embedding Similarity Network $g(\cdot)$:}

We introduce an embedding similarity network to allow the model to learn good representations with a strong correlation to the relevant LCC embeddings. $\hat z$ and $E$ are compared using the cosine similarity function:
\begin{equation}
    \label{eq:cosine}
    % c_i = \texttt{sim}_i(\hat{z_i}, E)= \|\hat{z_i}\|_{2} \times \|E\|_{2}
        c_i = \texttt{sim}_i(\hat{z_i}, E)= \frac{\hat{z_i}\cdot E}{\|\hat{z_i}\| \|E\|}
\end{equation}
where $i$ is the temporal index of $\hat{z}$ to produce the similarity score $c_i\in\left[-1,1\right]^{V' \times M}$. The resulting score is calculated by taking the average cosine similarity plus maximum cosine similarity across variations $M$ where $\hat{c}_i\in \mathbb{R}^{V'}$.
This is computed for each index in $T'$ such that $\hat{c}\in \mathbb{R}^{T' \times V'}$.
Using $\hat{c}$ we apply a temperature scalar $\tau$ followed by a softmax function as:
\begin{equation}
    q = g(z,E)=\texttt{softmax}(\hat{c} / \tau) 
\end{equation}
where $q \in \mathbb{R}^{T'\times V'}$.
% Overall our Vocabulary Similarity Network is as follows:
% \begin{equation}
%     q  = g(z,E) = \texttt{softmax}(\frac{\texttt{sim}(\hat{z}, E)}{\tau})
% \end{equation}
Our motivation is for the model to explicitly learn representations for the sign vocabulary for each time segment $T'$. 

\paragraph{Sign Recognition Head $r(\cdot)$:}

The network $g$ requires the label associated to each time segment $T'$, which are not available in sign recognition datasets. Sign recognition datasets only provide information that a sign exists within the given video clip. Isolated sign recognition datasets typically have signers in the resting pose which are irrelevant for the sign label.
We, therefore, create a sign recognition head $r$ with the assumption that given a sufficiently large temporal window $T$ for a given sequence $x$, if we know that the sign $v_j$ from vocabulary $V$ exist somewhere in the sequence $x$, then we can set the label $v_j$ to our target value. 

For our model to learn the existence of a sign we need a function $r(\cdot)$ that takes $q\in \mathbb{R}^{T'\times V'}$ and outputs $\hat{q} \in [0,1]^{V}$.

More formally, our objective for $r$ is to output $\hat{q}_j = 1$ when the target sign corresponds to the $j^{th}$ element in the vocabulary, and if not output $\hat{q}_j = 0$. To achieve this, we formulate $r$ as:
\begin{equation}
    \label{eq:recog_head}
    \hat{q}= r(q) = \left[ 
    \frac{\sum_{t=0}^{T'}{q_{jt}}}{\sum_{t=0}^{T'}\sum_{k=1}^{V}{q_{kt}}} \right]^{j: (1..V)}
\end{equation}

The output is then a $V$ length vector with values between 0 and 1, which indicate the existence of a sign from our given target vocabulary.

\subsection{Learning Objective}

\paragraph{Contrastive Recognition Loss $L_{rec}$:}

Unlike previous approaches which apply global average pooling from \cref{eq:normal}, our approach instead learns to localise the sign within the sequence using our weakly supervised contrastive recognition loss. 

From \cref{eq:recog_head}, $r$ allows the model to learn the existence of the sign in the given sequence since we only take the sum across the given target vocabulary $[1,...,V]$ and not the extended vocabulary $[(V+1),...,V']$. We apply a binary cross entropy loss for the contrastive recognition loss $L_{rec}$, where the target is the one-hot encoded vector of the label associated with the sign sequence.
This allows our model to bring matching signs closer together within the embedding space while pushing dissimilar signs further apart.

\paragraph{Conceptual Similarity Loss $L_{concept}$:}

Since sign language is a visual language, there are many signs that look very similar or the same, with small discriminating factors. For example, stomach and abdomen are signed in almost identical ways in the WLASL dataset. Therefore the representations should be similar. Generally speaking, when this is the case, the semantic similarity of the spoken word or gloss should also be similar. 

We introduce a new method to integrate sign language with spoken languages.
We use fastText embeddings \cite{bojanowski2017enriching} of the glosses and learn visual embeddings which learn the correlation between sign embeddings and spoken language embeddings from the target vocabulary. 
We propose a conceptual similarity loss $L_{concept}$, where we take the cosine similarity between the fastText embeddings $F$ for the target vocabulary to create a similarity matrix $S_F\in [-1,1]^{V\times V}$. The matrix is used to provide the model with a measure of the similarity of words.
We then repeat the process with the LCC embeddings $E_{1..V}$ to create the visual embedding cosine similarity $S_E\in [-1,1]^{V\times V}$.
A Mean Squared Error loss minimises the distance between $S_F$ and $S_E$. 
This allows our sign embeddings to learn similar linguistic embedding distributions as the word embeddings.

\paragraph{Combined Loss:}
We apply our learning objectives simultaneously during training. This allows the model to learn visual-linguistic representations and automatically localise signs within the given sequence. We apply a weighted loss $L_{rec}$ and $L_{concept}$ to create our LCC loss $\mathcal{L}$:
\begin{equation}
    \label{eq:learning_obj}
    \mathcal{L}=\alpha * L_{concept} + \beta * L_{rec}    
\end{equation}

\subsection{Drop Feature Mask}
\label{dropmask}
In the standard classification task, cross entropy loss is used with dropout before the final fully connected layer to improve the model's performance \cite{srivastava2014dropout}. Since we measure cosine similarity between features, we introduce drop feature masking to our representations $\hat z$ and $E$ where we zero out a portion of the same index channels in dimension $C$ for both $\hat z$ and $E$.
We also apply this at the temporal level which randomly zeros out the features at certain time segments.
This technique allows our model to create a better embedding representation by utilising more of the feature channels.

\subsection{Multi-channel Learning}
\label{sec:multi_channel}
\begin{figure}[h]
    \centering
    \includegraphics[width=\columnwidth]{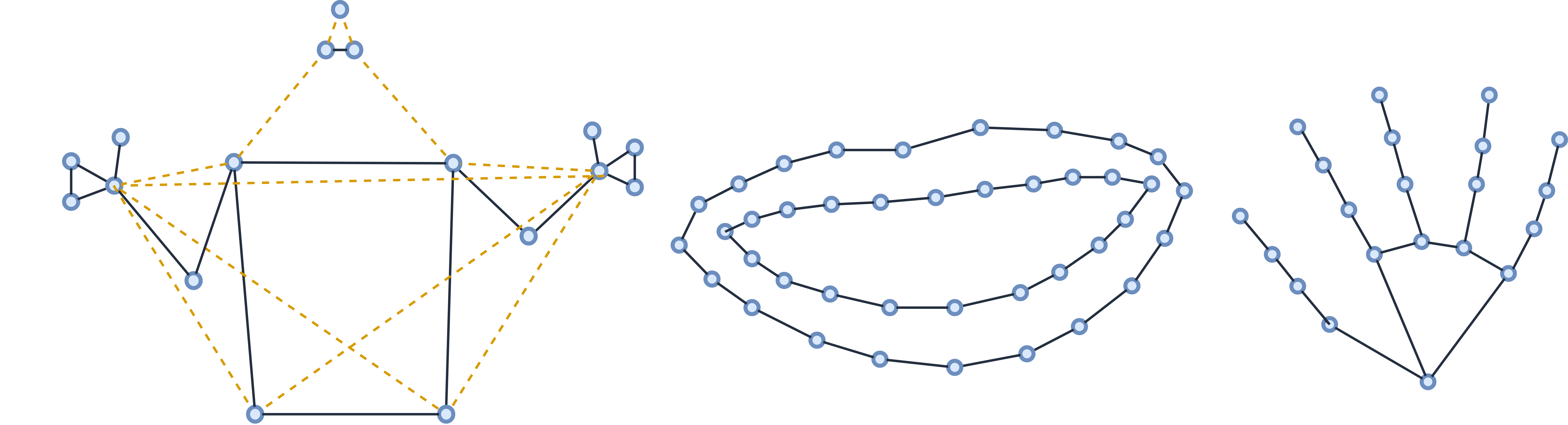}
    \caption{Keypoints extracted from Mediapipe used as the input into the different channels. Additional keypoint links shown by the dashed orange lines were added to allow the model to learn better hand-to-body interactions.}
    \label{fig:keypoint}
\end{figure}

We use a skeleton-based GCN as our backbone model. 
Skeleton representations, or keypoints, naturally provide a high degree of person independence by discarding irrelevant information, such as background, clothing or appearance. 
Jiang \etal highlights the issues of using a large number of keypoints and employs a graph reduction technique to significantly improve performance on sign recognition \cite{jiang2021sign}.
We use an alternative approach where different sign channels (body, hands and face) are input into separate GCN models. This allows our model's graph structure to remain small and gives it the ability to learn from each sign channel separately before fusing the sign features together.
We use Mediapipe Holistic \cite{lugaresi2019mediapipe} to extract the keypoints. We extract 4 different channels which include two hands, of 21 keypoints each, 40 keypoints representing the mouth and 17 keypoints for the body pose.
For the links between keypoints we use the default human skeletal connections with additional links on the body for a better hand-to-body interaction as shown in Figure~\ref{fig:keypoint}.

We create three separate GCN models and apply our learning objective from \cref{eq:learning_obj} on the output of the hands (where each hand has a shared backbone GCN and average the output hand features for $\hat{z}$), mouthing and pose as $\mathcal{L}_{h}, \mathcal{L}_m$ and $\mathcal{L}_{p}$. 
We also concatenate the $\hat{z}$ feature outputs across all GCNs outputs to create a global representation and use a fully connected layer to reduce the dimensions back to the original dimensionality and apply the same learning objective ($\mathcal{L}_{global}$) with a separate global LCC network head (with the LCC embedding, embedding similarity network and sign recognition head).
The final learning objective is:
\begin{equation}
     \mathcal{L}_{overall} = \mathcal{L}_{global}+\mathcal{L}_{h}+\mathcal{L}_m+\mathcal{L}_{p}
\end{equation}
At inference, the predicted label is selected based on the index of the largest value of $\hat{q}$ from the global LCC network head.

\section{Experiments}

\subsection{Datasets}

% \paragraph{Datasets.}
We evaluate our approach on two different sign recognition tasks: isolated sign recognition and the recognition of a co-articulated sign in continuous footage.

\textbf{WLASL2000} is a large-scale video dataset for American Sign Language (ASL) recognition. It is a challenging isolated sign recognition dataset as it was collected from unconstrained records with a large vocabulary of 2000 unique signs. Additional challenges are the signer-independent setting with over 100 signers and limited examples for each sign. 

\textbf{BOBSL} is a large-scale co-articulated sign dataset for British Sign Language (BSL) obtained from broadcast videos. The recognition task on the BOBSL dataset poses new challenges for sign recognition as continuous co-articulated sign is typically signed faster than in an isolated setting. The dataset contains 2281 sign classes obtained by using automatic spotting tools from mouthings and dictionary sources, which has the added difficulty of noisy labels and large class imbalances.

\subsection{Implementation Details}
\label{subsec:implementation}

We use the MS-G3N as our backbone GCN model due to its high performance on skeletal action recognition \cite{liu2020disentangling}.
For simplicity, we keep all model hyperparameters the same across each GCN. We keep the original hyperparameters from \cite{liu2020disentangling} but set the number of scales to 5 and use a window size of 5 with a dilation of 1 for the MS-G3D pathways.

% \subsection{Training Settings}
We use a sequence length of 64 and 16 for WLASL2000 and BOBSL, respectively. During training, rotation, scaling and shifting are used as keypoint data augmentation. 
We set the size of $V'$ to be $V+10$ with $\alpha$ of $5.0$ and $\beta$ of $10.0$ for the scaling factors of $L_{concept}$ and $L_{rec}$, respectively.
We train the model with a batch size of 64 with a learning rate of 0.0012.
On the WLASL dataset, we schedule our learning rate with a warm-up of 10 epochs and decay with a cosine policy to match the training strategy of ~\cite{jiang2021skeleton} on the WLASL dataset, but train our model for 100 epochs instead of 200.
We use a multi-step learning rate scheduler for the BOBSL dataset by reducing the learning rate by 0.1 at 10 and 20 epochs and train our model for 25 epochs in total.

The model is optimized using the Adam optimizer \cite{kingma2014adam} with a weight decay of 0.0001, where we select the model which has the best accuracy on the validation set for evaluation on the test set.

\subsection{Evaluation Protocol}
We evaluate our models on the top-1 and top-5 per-instance classification accuracy as well as the top-1 and top-5 per-class accuracy.
Our proposed training strategy is compared to our model trained with cross entropy loss approach, using global average pooling and a classification layer, to directly analyse the impact of our approach.
Then, we compare our approach to state-of-the-art keypoints results on both WLASL2000 and BOBSL datasets. We also demonstrate the effectiveness of our framework to RGB-based sign recognition models.

\subsection{Results on Isolated Sign Recognition}

\begin{table}[ht]
    \centering
    \begin{tabular}[width=0.5\textwidth]{c c c c c}
        \toprule
        \multirow{2}{*}{Models} & \multicolumn{2}{c}{Instance Acc.} & \multicolumn{2}{c}{Class Acc.}\\
        \cmidrule{2-4} \cmidrule{5-5}
        % \cmidrule{2-4} \cmidrule{5-5} \\
        {} & top-1 & top-5 &  top-1  & top-5 \\
        \midrule
        SignBERT (H+P) \cite{hu2021signbert} & 47.46 & 83.32 & 45.17 & 82.32 \\
        BEST (Keypoint) \cite{zhao2023best} & 46.25 & 79.33 & 43.52 & 77.65 \\
        % \midrule
        SL-GCN \cite{jiang2021skeleton}&51.50& 84.94 &48.87& 84.02\\
        \midrule
        Ours (CE) & 51.95 & 82.52 & 48.89 & 81.15 \\
        Ours (LCC) & \textbf{59.38} & \textbf{89.82} & \textbf{56.57} & \textbf{88.90} \\
        % Ours (Weighted) & \textbf{59.49} & 89.96 & 56.59 & 89.06 \\
        \bottomrule
    \end{tabular}
    % \vspace{1cm}
    \caption{Per-instance and per-class accuracy on the WLASL2000 test set for the keypoint modality models. CE and LCC correspond to our models trained with cross entropy and LCC loss, respectively.}
    \label{tab:wlasl_accuracy}
\end{table}

% \vspace{-1.0cm}
\paragraph{Comparisons to baseline:}
We compare our proposed loss to cross entropy loss used in previous methods on the WLASL2000 dataset. In Table~\ref{tab:wlasl_accuracy}, we find that our model trained with cross entropy loss achieves similar performance to the SL-GCN model. 
The addition of our loss significantly improves the test accuracy by 7.43\% top-1 instance accuracy. Furthermore, the proposed approach has the additional benefit of providing localisation of the target sign as shown in Section~\ref{qualitativeanalysis}.

\paragraph{Comparison to state-of-the-art:}
In Table~\ref{tab:wlasl_accuracy}, we find that our multi-stream keypoint model is able to significantly outperform other keypoint modality models.
Our multi-stream keypoint model is able to outperform the previous multi-stream keypoint model SL-GCN by improving the top-1 instance accuracy by 7.88\%. 
While we are aware of Bidirectional Skeleton-Based Graph Convolutional Networks \cite{dafnis2022bidirectional}, we are unable to directly compare against it as the approach was evaluated on an unreleased modified WLASL dataset with 1449 lexical signs and contains additional preprocessing to detect the sign's start and stopping frames. Our model has the additional benefit of automatically detecting the sign's start and stopping frames.

We find that our keypoint modality results also achieve competitive results with a multi-modal ensemble such as SAM-SLR v1 and v2 \cite{jiang2021skeleton,jiang2021sign} which make use of keypoints, features, RGB frames and RGB flow as input modalities to the multiple models to achieve 59.39\% and 91.48\% top-1 and top-5 per-instance accuracy respectively. 

Further analysis of the individual streams of joint, bone, joint motion and bone motion in Table~\ref{tab:keypoint_wlasl_accuracy} shows significant individual stream performance improvements compared to SL-GCN. Our joint-based model is also able to outperform all previous single model results.

\begin{table}[ht]
    \centering
    \begin{tabular}[width=0.5\textwidth]{c c c c}
        \toprule
        \multirow{2}{*}{Stream} & \multirow{2}{*}{Models}& \multicolumn{2}{c}{Instance Acc.}\\
        \cmidrule{3-4} \cmidrule{4-4}
        % \cmidrule{2-4} \cmidrule{5-5} \\
        {} &{} & top-1 & top-5  \\
        \midrule
        \multirow{2}{*}{Joint} & SL-GCN  & 45.61 & 77.79\\
         & Ours (LCC)  & 55.52 &  86.76\\
        \midrule
        \multirow{2}{*}{Bone} & SL-GCN  & 43.27 & 75.58\\
         & Ours (LCC) & 54.37 & 85.78\\
        \midrule
         \multirow{2}{*}{Joint Motion} & SL-GCN  & 27.23 & 56.73\\
         & Ours (LCC) & 46.56 & 78.76 \\
        \midrule
         \multirow{2}{*}{Bone Motion} & SL-GCN  & 31.26 & 60.35\\
         & Ours (LCC) & 46.66 & 76.37 \\
        \midrule
        \midrule
        \multirow{2}{*}{Multi-stream} & SL-GCN  & 51.50 & 84.94\\
         & Ours (LCC) & \textbf{59.38} & \textbf{89.82} \\
        \bottomrule
    \end{tabular}
    \caption{Comparison between our LCC approach and SL-GCN on individual keypoint streams on the WLASL test set.}
    \label{tab:keypoint_wlasl_accuracy}
\end{table}

\subsection{Results on Sign Recognition in continuous footage}

\begin{table}[ht]
    \centering
    \begin{tabular}[width=0.5\textwidth]{c c c c c}
        \toprule
        \multirow{2}{*}{Models} & \multicolumn{2}{c}{Instance Acc.} & \multicolumn{2}{c}{Class Acc.}\\
        \cmidrule{2-4} \cmidrule{5-5}
        % \cmidrule{2-4} \cmidrule{5-5} \\
        {} & top-1 & top-5 &  top-1  & top-5 \\
        \midrule
        2D Pose $\to$ Sign \cite{albanie2021bbc} & 61.8 & 82.1 & 30.6 & 56.6 \\
        \midrule        
        Ours (CE) & 67.8 & 87.0 & 34.9 & 60.9 \\
        Ours (LCC) & \textbf{71.7} & \textbf{89.3} & \textbf{37.3} & \textbf{64.5} \\
        \bottomrule
    \end{tabular}
    \caption{Comparison of accuracy on the BOBSL test set. CE: Our model trained with Cross Entropy loss, LCC: Our model trained with LCC loss.}
    \label{tab:bobsl_accuracy}
\end{table}

\paragraph{Comparison to baseline:}

In Table~\ref{tab:bobsl_accuracy}, we show that our model trained with our loss is able to improve multi-stream model performance by almost 4\% on the top-1 instance accuracy compared to cross entropy loss on the BOBSL recognition task.

Further analysis of the keypoint modality on the individual streams of joint, bone, joint motion and bone motion in Table~\ref{tab:keypoint_bobsl_accuracy} shows similar improvements compared to cross entropy loss on individual streams. Our joint based model is able to produce the best individual stream results.

\begin{table}[ht]
    \centering
    \begin{tabular}[width=0.5\textwidth]{c c c c}
        \toprule
        \multirow{2}{*}{Stream} & \multirow{2}{*}{Loss}& \multicolumn{2}{c}{Instance Acc.}\\
        \cmidrule{3-4} \cmidrule{4-4}
        % \cmidrule{2-4} \cmidrule{5-5} \\
        {} &{} & top-1 & top-5  \\
        \midrule
        \multirow{2}{*}{Joint} & CE  & 66.28 & 86.02\\
         & LCC   & 70.89 & 88.83\\
        \midrule
        \multirow{2}{*}{Bone} & CE & 66.02 & 85.71\\
         & LCC   & 70.54 & 88.61\\
        \midrule
         \multirow{2}{*}{Joint Motion} & CE  & 61.05 & 81.60\\
         & LCC    & 67.49 & 86.18\\
        \midrule
         \multirow{2}{*}{Bone Motion} & CE  & 61.02 & 81.80\\
         & LCC  & 66.93 & 85.68\\
        \midrule
        \midrule
        \multirow{2}{*}{Multi-stream} & CE  & 67.75 & 86.99 \\
         & LCC   & \textbf{71.72} & \textbf{89.32}\\
        \bottomrule
    \end{tabular}
    \caption{Comparison between our model trained with cross entropy loss (CE) versus our LCC loss on the BOBSL test set.}
    \label{tab:keypoint_bobsl_accuracy}
\end{table}

\paragraph{Comparison to state-of-the-art:}
In Table~\ref{tab:bobsl_accuracy}, we show that our approach is able to outperform the previous keypoint based model (2D Pose$\to$Sign) by almost 10\% top-1 accuracy. All of our individual stream results are able to outperform 2D Pose$\to$Sign by more than 5\% top-1 accuracy.

\begin{figure*}[ht]
    \centering
    \includegraphics[width=0.95\textwidth]{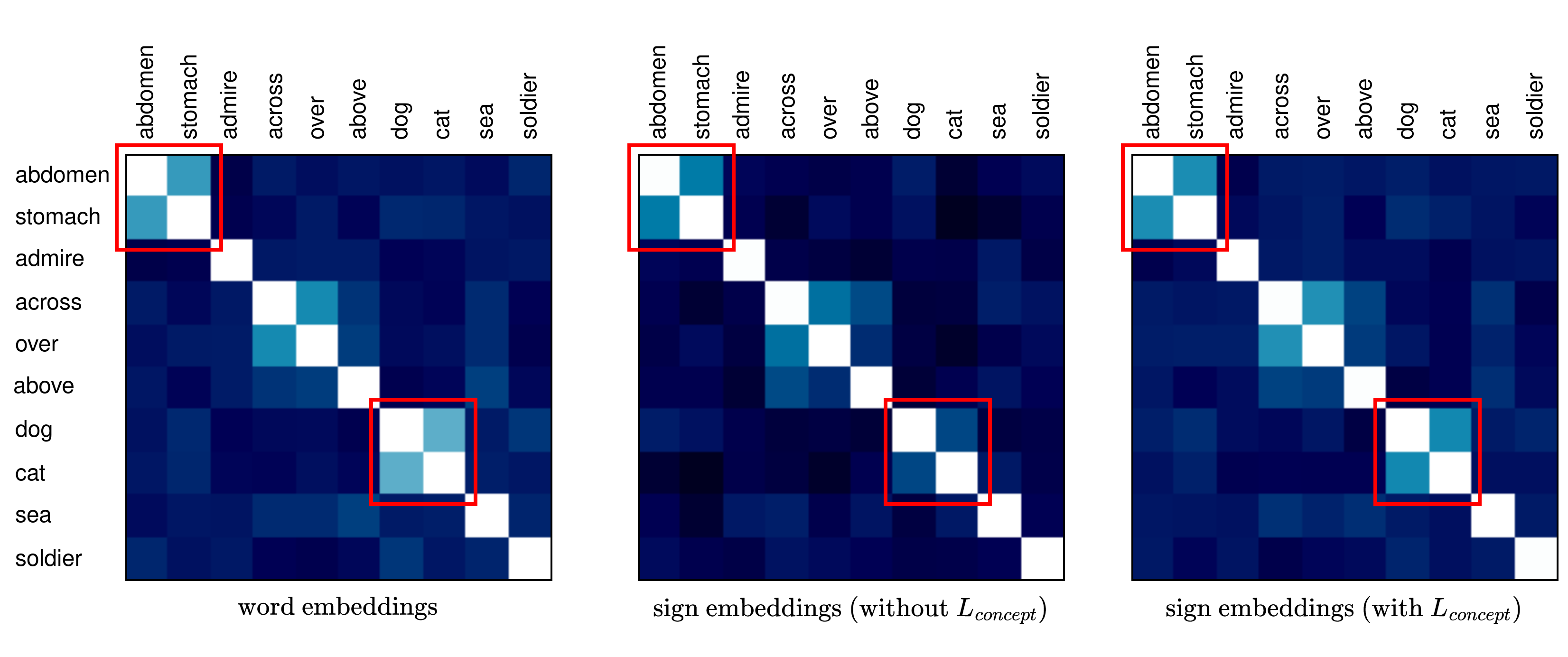}
    \caption{Comparison of similarity matrices for the word embeddings from fastText (left), sign embeddings from a model trained without conceptual similarity loss (middle) and sign embeddings trained with conceptual similarity loss (right). Red regions highlight that the similarity matrix trained with the conceptual similarity has similar linguistic embedding distributions to the word embeddings.}
    \label{fig:sim_matrix}
\end{figure*}

\subsection{Ablation Study}

We perform our ablation study on the WLASL validation dataset using the joint-based stream for our model. 

\paragraph{Impact of temperature:}
The temperature $\tau$ plays an important part in the model's performance. We find that when the temperature is too high, it has a negative impact on the model's performance. In Table~\ref{tab:temperature_table}, $\tau=0.1$ gives the strongest performance improvements.

\begin{table}[ht]
    \centering
    \begin{tabular}{c c c}
        \toprule
        \multirow{2}{*}{$\tau$} & \multicolumn{2}{c}{Instance Acc.}\\
        \cmidrule{2-3}
        % \cmidrule{2-4} \cmidrule{5-5} \\
        {} & top-1 & top-5  \\
        \midrule
        0.5 & 0.1938 & 0.4816 \\
        0.2 & 0.3897 & 0.7181  \\
        \textbf{0.1} & \textbf{0.5181} & \textbf{0.8307}  \\
        0.05 & 0.5023 & 0.8159 \\
        \midrule
    \end{tabular}
    \caption{Table demonstrating the impact of the temperature value has on the WLASL validation accuracy}
    \label{tab:temperature_table}
\end{table}

\paragraph{Impact of $L_{concept}$:}
Due to the recognition loss $L_{rec}$ having a contrastive objective to match representations from the model to the target vocabulary embedding, the model tends to push other representations further apart. This is detrimental in cases where signs are visually similar. Our Conceptual Similarity Loss $L_{concept}$ attempts to alleviate these issues by creating visual embeddings which match the distances between words from word embeddings. In Table~\ref{tab:aux_table}, we show that the inclusion of $L_{concept}$ greatly improves the accuracy.
\begin{table}[ht]
    \centering
    \begin{tabular}{c c c c}
        \toprule
        $L_{rec} $ & $L_{concept}$ & \multicolumn{2}{c}{Instance Acc.}\\
        \cmidrule{3-4}
        % \cmidrule{2-4} \cmidrule{5-5} \\
        $\beta$ & $\alpha$  & top-1 & top-5  \\
        \midrule
        10.0 & 0.0 & 52.76 & 82.79 \\
        10.0  & 1.0 & 53.32 & 84.53  \\
        10.0  & 5.0 & \textbf{54.29} & \textbf{84.60}  \\
        10.0  & 10.0 & 54.11 & 83.96  \\
        \midrule
    \end{tabular}
    \caption{Table demonstrating the importance and impact that the cosine similarity loss has on the WLASL validation accuracy}
    \label{tab:aux_table}
\end{table}

In Figure~\ref{fig:sim_matrix}, we show the impact the conceptual similarity loss has on the embeddings by computing the cosine similarity between our sign embeddings. As shown, without the similarity loss the model tends to learn a similar structure to the word embedding. The conceptual similarity loss is able to regularise features in our embedding space to match the distribution of the associated word embedding space. For example, the signs for stomach and abdomen in Figure~\ref{fig:sim_matrix} have higher similarity scores (brighter) with a model trained with $L_{concept}$ (right) than without (middle).

\paragraph{Impact of drop feature masking:}
We introduce drop feature masking on the embedding features for both the model outputs and vocabulary embedding to create representations that reduce overfitting and over-reliance on certain feature channels. In Table~\ref{tab:dropmask_table}, we show that the inclusion of drop masking improves performance.
\begin{table}[ht]
    \centering
    \begin{tabular}{c c c}
        \toprule
        \multirow{2}{*}{dfm} & \multicolumn{2}{c}{Instance Acc.}\\
        \cmidrule{2-3}
        % \cmidrule{2-4} \cmidrule{5-5} \\
        {}   & top-1 & top-5  \\
        \midrule
        $-$ & 54.52 & 85.32 \\
        $\checkmark$   & \textbf{56.66} & \textbf{87.16} \\
        \midrule
    \end{tabular}
    \caption{Table demonstrating the impact of drop feature masking on WLASL validation results.}
    \label{tab:dropmask_table}
\end{table}

\paragraph{Impact of backbone:}

As an additional study, we evaluate our approach using an RGB backbone model to demonstrate that our framework is input-agnostic. 
A limitation of multi-channel RGB inputs is the heavy computational requirements. Separate video crops for each sign channel are more memory intensive compared to the keypoint based approach, we therefore use the I3D model with full frames as input.
Previous methods have shown that transfer learning of models on other sign language datasets improves sign recognition performance \cite{albanie2020bsl,jiang2021sign}. 
We, therefore, use Inception I3D pretrained on Kinetic dataset \cite{kay2017kinetics} to directly evaluate the impact of our approach compared to cross entropy loss without any pretraining on external sign language datasets. We apply our learning objective from \cref{eq:learning_obj}.
For data augmentation during training, we first apply frame resizing to $256\times 256$ then random cropping of $224\times 224$ with random horizontal flipping and colour jitter.

\begin{table}[ht]
    \centering
    \begin{tabular}[width=0.5\textwidth]{c c c c c}
        \toprule
        \multirow{2}{*}{Models} & \multicolumn{2}{c}{Instance Acc.} & \multicolumn{2}{c}{Class Acc.}\\
        \cmidrule{2-4} \cmidrule{5-5}
        % \cmidrule{2-4} \cmidrule{5-5} \\
        {} & top-1 & top-5 &  top-1  & top-5 \\
        \midrule
        I3D (CE) & 41.38 & 74.98 & 38.93 & 73.94 \\
        I3D (LCC)  & \textbf{43.92} & \textbf{77.80} & \textbf{41.10} & \textbf{75.97} \\
        \bottomrule
    \end{tabular}
    \caption{Table demonstrating the difference between our LCC loss with a RGB backbone (LCC) versus cross entropy loss (CE) on the WLASL test set.}
    \label{tab:wlasl_rgb}
\end{table}

In Table~\ref{tab:wlasl_rgb}, we show that our proposed loss is able to improve performance compared to cross entropy loss. This experiment demonstrates that our model is capable of generalising to RGB models.

\subsection{Sign Localisation}

\label{qualitativeanalysis}
\begin{figure}[h]
    \centering
    \includegraphics[width=\columnwidth]{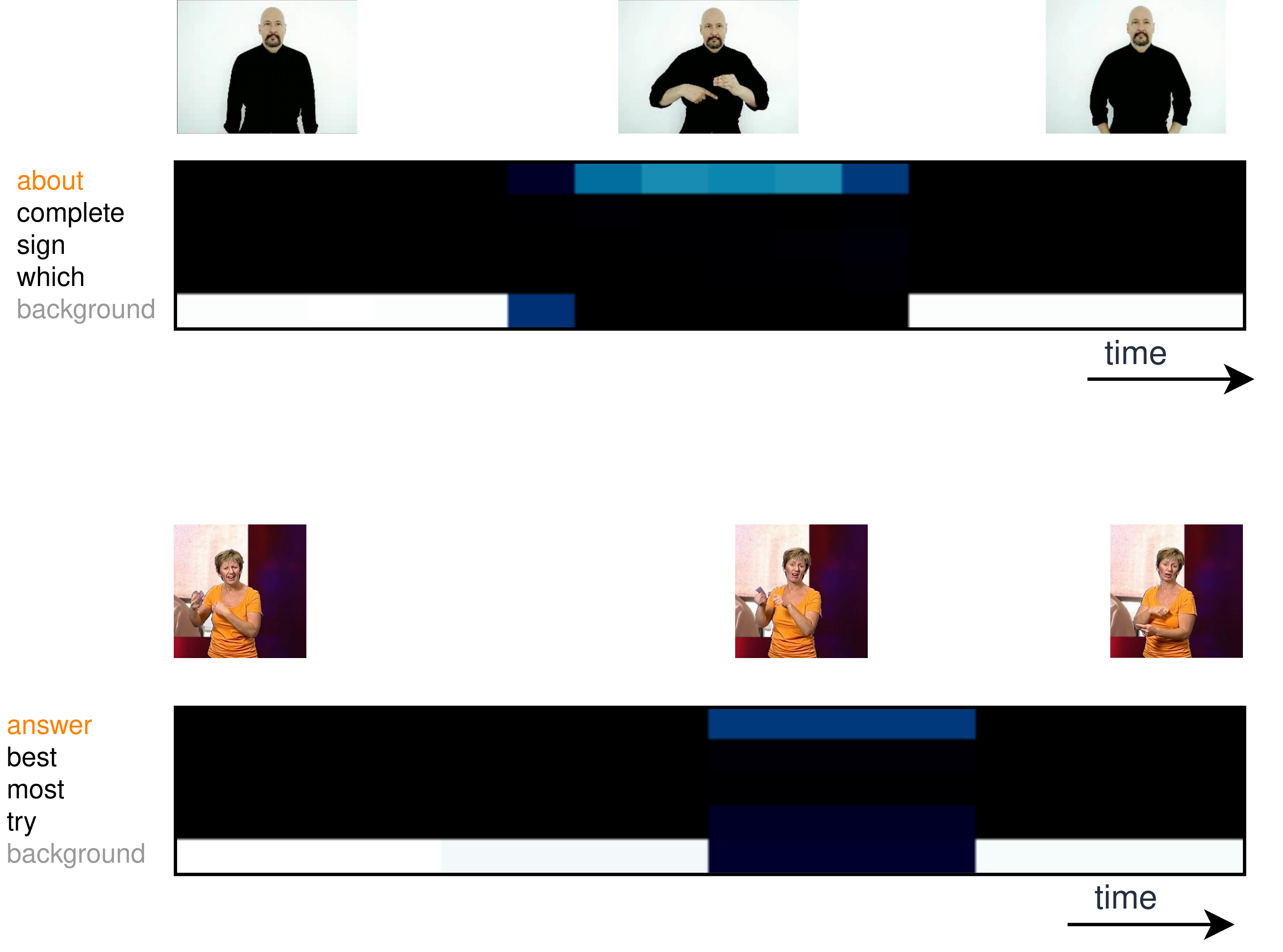}
    \caption{Localisation of identified signs, target word highlighted in orange, where temporal activations are shown in brighter colours. The background class is the sum of the probabilities across extended vocabulary.}
    \label{fig:localisation}
\end{figure}

Our model is able to temporally locate the target sign in a given sign sequence using the output value $q$. In Figure~\ref{fig:localisation}, we show the visualisation of the localisation of the signs. We find that in WLASL2000 there are many sequences with signers in their resting pose, which are irrelevant to the target vocabulary. Our model is able to identify those background segments. Similarly, for BOBSL, our model is able to identify when a sign is out of vocabulary and localise the target vocabulary signs.

\subsection{Limitations}

One of the limitations of the GCN model is its reliance on the accuracy of keypoint inputs. Past research has highlighted that existing pretrained pose estimators may fail in hand-to-hand or hand-to-face interactions \cite{moryossef2021evaluating}.

Secondly, our framework assumes that the conceptual similarity of the words often implies visual similarity in sign language. While we show experiments on the inclusion of the conceptual similarity loss that supports our hypothesis, it may not always be the case for all signs, for example on the signs of names and places.

\section{Conclusion}

In this work, we introduce Learnt Contrastive Concept embeddings framework, a novel training strategy to learn sign embeddings by employing a weakly supervised contrastive training pipeline that is able to learn sign embeddings from the sign recognition task. We demonstrate the effectiveness of our approach for the localisation of signs and how our framework can be used to improve results on sign recognition compared to cross entropy loss.
Our model is able to significantly outperform previous keypoint recognition results on both WLASL and BOBSL datasets. 
Our approach is able to utilise word embeddings to create sign embeddings that incorporate visual-linguistic features that will hopefully be useful for future work in sign language translation.
For future work, exploration of improving the framework for continuous sign recognition may be useful to solve as the next step for automatic sign language translation research.

\section*{Acknowledgements}
This work was supported by the EPSRC project ExTOL (EP/R03298X/1), SNSF project ’SMILE II’ (CRSII5 193686), European Union’s Horizon2020 programme (’EASIER’ grant agreement 101016982) and the Innosuisse IICT Flagship (PFFS-21-47). This work reflects only the authors view and the Commission is not responsible for any use that may be made of the information it contains

{\small
\bibliographystyle{ieee_fullname}
\bibliography{egbib}

\begin{thebibliography}{10}\itemsep=-1pt

\bibitem{albanie2020bsl}
Samuel Albanie, G{\"u}l Varol, Liliane Momeni, Triantafyllos Afouras, Joon~Son
  Chung, Neil Fox, and Andrew Zisserman.
\newblock Bsl-1k: Scaling up co-articulated sign language recognition using
  mouthing cues.
\newblock In {\em European Conference on Computer Vision}, pages 35--53.
  Springer, 2020.

\bibitem{albanie2021bbc}
Samuel Albanie, G{\"u}l Varol, Liliane Momeni, Hannah Bull, Triantafyllos
  Afouras, Himel Chowdhury, Neil Fox, Bencie Woll, Rob Cooper, Andrew
  McParland, et~al.
\newblock Bbc-oxford british sign language dataset.
\newblock {\em arXiv preprint arXiv:2111.03635}, 2021.

\bibitem{bilge2022towards}
Yunus~Can Bilge, Ramazan~Gokberk Cinbis, and Nazli Ikizler-Cinbis.
\newblock Towards zero-shot sign language recognition.
\newblock {\em IEEE Transactions on Pattern Analysis and Machine Intelligence},
  2022.

\bibitem{bojanowski2017enriching}
Piotr Bojanowski, Edouard Grave, Armand Joulin, and Tomas Mikolov.
\newblock Enriching word vectors with subword information.
\newblock {\em Transactions of the Association for Computational Linguistics},
  5:135--146, 2017.

\bibitem{camgoz2018neural}
Necati~Cihan Camgoz, Simon Hadfield, Oscar Koller, Hermann Ney, and Richard
  Bowden.
\newblock Neural sign language translation.
\newblock In {\em Proceedings of the IEEE conference on computer vision and
  pattern recognition}, pages 7784--7793, 2018.

\bibitem{camgoz2020multi}
Necati~Cihan Camgoz, Oscar Koller, Simon Hadfield, and Richard Bowden.
\newblock Multi-channel transformers for multi-articulatory sign language
  translation.
\newblock In {\em European Conference on Computer Vision}, pages 301--319.
  Springer, 2020.

\bibitem{camgoz2020sign}
Necati~Cihan Camgoz, Oscar Koller, Simon Hadfield, and Richard Bowden.
\newblock Sign language transformers: Joint end-to-end sign language
  recognition and translation.
\newblock In {\em Proceedings of the IEEE/CVF conference on computer vision and
  pattern recognition}, pages 10023--10033, 2020.

\bibitem{cao2019openpose}
Zhe Cao, Gines Hidalgo, Tomas Simon, Shih-En Wei, and Yaser Sheikh.
\newblock Openpose: realtime multi-person 2d pose estimation using part
  affinity fields.
\newblock {\em IEEE transactions on pattern analysis and machine intelligence},
  43(1):172--186, 2019.

\bibitem{carreira2017quo}
Joao Carreira and Andrew Zisserman.
\newblock Quo vadis, action recognition? a new model and the kinetics dataset.
\newblock In {\em proceedings of the IEEE Conference on Computer Vision and
  Pattern Recognition}, pages 6299--6308, 2017.

\bibitem{chen2020improved}
Xinlei Chen, Haoqi Fan, Ross Girshick, and Kaiming He.
\newblock Improved baselines with momentum contrastive learning.
\newblock {\em arXiv preprint arXiv:2003.04297}, 2020.

\bibitem{cihan2017subunets}
Necati Cihan~Camgoz, Simon Hadfield, Oscar Koller, and Richard Bowden.
\newblock Subunets: End-to-end hand shape and continuous sign language
  recognition.
\newblock In {\em Proceedings of the IEEE international conference on computer
  vision}, pages 3056--3065, 2017.

\bibitem{dafnis2022bidirectional}
Konstantinos~M Dafnis, Evgenia Chroni, Carol Neidle, and Dimitri Metaxas.
\newblock Bidirectional skeleton-based isolated sign recognition using graph
  convolutional networks.
\newblock In {\em Proceedings of the Thirteenth Language Resources and
  Evaluation Conference}, pages 7328--7338, 2022.

\bibitem{forster2014extensions}
Jens Forster, Christoph Schmidt, Oscar Koller, Martin Bellgardt, and Hermann
  Ney.
\newblock Extensions of the sign language recognition and translation corpus
  rwth-phoenix-weather.
\newblock In {\em LREC}, pages 1911--1916, 2014.

\bibitem{hanke2004hamnosys}
Thomas Hanke.
\newblock Hamnosys-representing sign language data in language resources and
  language processing contexts.
\newblock In {\em LREC}, volume~4, pages 1--6, 2004.

\bibitem{hao2021self}
Aiming Hao, Yuecong Min, and Xilin Chen.
\newblock Self-mutual distillation learning for continuous sign language
  recognition.
\newblock In {\em Proceedings of the IEEE/CVF International Conference on
  Computer Vision}, pages 11303--11312, 2021.

\bibitem{hu2021signbert}
Hezhen Hu, Weichao Zhao, Wengang Zhou, Yuechen Wang, and Houqiang Li.
\newblock Signbert: pre-training of hand-model-aware representation for sign
  language recognition.
\newblock In {\em Proceedings of the IEEE/CVF International Conference on
  Computer Vision}, pages 11087--11096, 2021.

\bibitem{jiang2021sign}
Songyao Jiang, Bin Sun, Lichen Wang, Yue Bai, Kunpeng Li, and Yun Fu.
\newblock Sign language recognition via skeleton-aware multi-model ensemble.
\newblock {\em arXiv preprint arXiv:2110.06161}, 2021.

\bibitem{jiang2021skeleton}
Songyao Jiang, Bin Sun, Lichen Wang, Yue Bai, Kunpeng Li, and Yun Fu.
\newblock Skeleton aware multi-modal sign language recognition.
\newblock In {\em Proceedings of the IEEE/CVF Conference on Computer Vision and
  Pattern Recognition}, pages 3413--3423, 2021.

\bibitem{joze2018ms}
Hamid Reza~Vaezi Joze and Oscar Koller.
\newblock Ms-asl: A large-scale data set and benchmark for understanding
  american sign language.
\newblock {\em arXiv preprint arXiv:1812.01053}, 2018.

\bibitem{kay2017kinetics}
Will Kay, Joao Carreira, Karen Simonyan, Brian Zhang, Chloe Hillier, Sudheendra
  Vijayanarasimhan, Fabio Viola, Tim Green, Trevor Back, Paul Natsev, et~al.
\newblock The kinetics human action video dataset.
\newblock {\em arXiv preprint arXiv:1705.06950}, 2017.

\bibitem{khosla2020supervised}
Prannay Khosla, Piotr Teterwak, Chen Wang, Aaron Sarna, Yonglong Tian, Phillip
  Isola, Aaron Maschinot, Ce Liu, and Dilip Krishnan.
\newblock Supervised contrastive learning.
\newblock {\em Advances in Neural Information Processing Systems},
  33:18661--18673, 2020.

\bibitem{kingma2014adam}
Diederik~P Kingma and Jimmy Ba.
\newblock Adam: A method for stochastic optimization.
\newblock {\em arXiv preprint arXiv:1412.6980}, 2014.

\bibitem{koller2019weakly}
Oscar Koller, Necati~Cihan Camgoz, Hermann Ney, and Richard Bowden.
\newblock Weakly supervised learning with multi-stream cnn-lstm-hmms to
  discover sequential parallelism in sign language videos.
\newblock {\em IEEE transactions on pattern analysis and machine intelligence},
  42(9):2306--2320, 2019.

\bibitem{koller2015deep}
Oscar Koller, Hermann Ney, and Richard Bowden.
\newblock Deep learning of mouth shapes for sign language.
\newblock In {\em Proceedings of the IEEE International Conference on Computer
  Vision Workshops}, pages 85--91, 2015.

\bibitem{li2020word}
Dongxu Li, Cristian Rodriguez, Xin Yu, and Hongdong Li.
\newblock Word-level deep sign language recognition from video: A new
  large-scale dataset and methods comparison.
\newblock In {\em Proceedings of the IEEE/CVF winter conference on applications
  of computer vision}, pages 1459--1469, 2020.

\bibitem{liu2020disentangling}
Ziyu Liu, Hongwen Zhang, Zhenghao Chen, Zhiyong Wang, and Wanli Ouyang.
\newblock Disentangling and unifying graph convolutions for skeleton-based
  action recognition.
\newblock In {\em Proceedings of the IEEE/CVF conference on computer vision and
  pattern recognition}, pages 143--152, 2020.

\bibitem{lugaresi2019mediapipe}
Camillo Lugaresi, Jiuqiang Tang, Hadon Nash, Chris McClanahan, Esha Uboweja,
  Michael Hays, Fan Zhang, Chuo-Ling Chang, Ming~Guang Yong, Juhyun Lee, et~al.
\newblock Mediapipe: A framework for building perception pipelines.
\newblock {\em arXiv preprint arXiv:1906.08172}, 2019.

\bibitem{mikolov2013efficient}
Tomas Mikolov, Kai Chen, Greg Corrado, and Jeffrey Dean.
\newblock Efficient estimation of word representations in vector space.
\newblock {\em arXiv preprint arXiv:1301.3781}, 2013.

\bibitem{min2021visual}
Yuecong Min, Aiming Hao, Xiujuan Chai, and Xilin Chen.
\newblock Visual alignment constraint for continuous sign language recognition.
\newblock In {\em Proceedings of the IEEE/CVF International Conference on
  Computer Vision}, pages 11542--11551, 2021.

\bibitem{momeni2022automatic}
Liliane Momeni, Hannah Bull, KR Prajwal, Samuel Albanie, G{\"u}l Varol, and
  Andrew Zisserman.
\newblock Automatic dense annotation of large-vocabulary sign language videos.
\newblock In {\em ECCV 2022: 17th European Conference, Tel Aviv, Israel,
  October 23--27, 2022, Proceedings, Part XXXV}, pages 671--690. Springer,
  2022.

\bibitem{moryossef2021evaluating}
Amit Moryossef, Ioannis Tsochantaridis, Joe Dinn, Necati~Cihan Camgoz, Richard
  Bowden, Tao Jiang, Annette Rios, Mathias Muller, and Sarah Ebling.
\newblock Evaluating the immediate applicability of pose estimation for sign
  language recognition.
\newblock In {\em Proceedings of the IEEE/CVF Conference on Computer Vision and
  Pattern Recognition}, pages 3434--3440, 2021.

\bibitem{pennington2014glove}
Jeffrey Pennington, Richard Socher, and Christopher~D Manning.
\newblock Glove: Global vectors for word representation.
\newblock In {\em Proceedings of the 2014 conference on empirical methods in
  natural language processing (EMNLP)}, pages 1532--1543, 2014.

\bibitem{schroff2015facenet}
Florian Schroff, Dmitry Kalenichenko, and James Philbin.
\newblock Facenet: A unified embedding for face recognition and clustering.
\newblock In {\em Proceedings of the IEEE conference on computer vision and
  pattern recognition}, pages 815--823, 2015.

\bibitem{sincan2020autsl}
Ozge~Mercanoglu Sincan and Hacer~Yalim Keles.
\newblock Autsl: A large scale multi-modal turkish sign language dataset and
  baseline methods.
\newblock {\em IEEE Access}, 8:181340--181355, 2020.

\bibitem{srivastava2014dropout}
Nitish Srivastava, Geoffrey Hinton, Alex Krizhevsky, Ilya Sutskever, and Ruslan
  Salakhutdinov.
\newblock Dropout: a simple way to prevent neural networks from overfitting.
\newblock {\em The journal of machine learning research}, 15(1):1929--1958,
  2014.

\bibitem{stokoe1980sign}
William~C Stokoe.
\newblock Sign language structure.
\newblock {\em Annual review of anthropology}, pages 365--390, 1980.

\bibitem{sutton2009signwriting}
Valerie Sutton.
\newblock Signwriting.
\newblock {\em Sl: sn}, page~9, 2009.

\bibitem{tran2018closer}
Du Tran, Heng Wang, Lorenzo Torresani, Jamie Ray, Yann LeCun, and Manohar
  Paluri.
\newblock A closer look at spatiotemporal convolutions for action recognition.
\newblock In {\em Proceedings of the IEEE conference on Computer Vision and
  Pattern Recognition}, pages 6450--6459, 2018.

\bibitem{yan2018spatial}
Sijie Yan, Yuanjun Xiong, and Dahua Lin.
\newblock Spatial temporal graph convolutional networks for skeleton-based
  action recognition.
\newblock In {\em Thirty-second AAAI conference on artificial intelligence},
  2018.

\bibitem{zhang2020mediapipe}
Fan Zhang, Valentin Bazarevsky, Andrey Vakunov, Andrei Tkachenka, George Sung,
  Chuo-Ling Chang, and Matthias Grundmann.
\newblock Mediapipe hands: On-device real-time hand tracking.
\newblock {\em arXiv preprint arXiv:2006.10214}, 2020.

\bibitem{zhao2023best}
Weichao Zhao, Hezhen Hu, Wengang Zhou, Jiaxin Shi, and Houqiang Li.
\newblock Best: Bert pre-training for sign language recognition with coupling
  tokenization.
\newblock {\em arXiv preprint arXiv:2302.05075}, 2023.

\end{thebibliography}
}

\end{document}